# Evolved Explainable Classifications for Lymph Node Metastases

Iam Palatnik de Sousa, Marley M B R Vellasco, *Senior Member, IEEE,* and Eduardo Costa da Silva, *Member, IEEE*

***Abstract*— A novel evolutionary approach for Explainable Artificial Intelligence is presented: the "Evolved Explanations" model (EvEx). This methodology consists in combining Local Interpretable Model Agnostic Explanations (LIME) with Multi-Objective Genetic Algorithms to allow for automated segmentation parameter tuning in image classification tasks. In this case, the dataset studied is Patch-Camelyon, comprised of patches from pathology whole slide images. A publicly available Convolutional Neural Network (CNN) was trained on this dataset to provide a binary classification for presence/absence of lymph node metastatic tissue. In turn, the classifications are explained by means of evolving segmentations, seeking to optimize three evaluation goals simultaneously. The final explanation is computed as the mean of all explanations generated by Pareto front individuals, evolved by the developed genetic algorithm. To enhance reproducibility and traceability of the explanations, each of them was generated from several different seeds, randomly chosen. The observed results show remarkable agreement between different seeds. Despite the stochastic nature of LIME explanations, regions of high explanation weights proved to have good agreement in the heat maps, as computed by pixel-wise relative standard deviations. The found heat maps coincide with expert medical segmentations, which demonstrates that this methodology can find high quality explanations (according to the evaluation metrics), with the novel advantage of automated parameter fine tuning. These results give additional insight into the inner workings of neural network black box decision making for medical data.**

## I. Introduction

THE medical domain has seen an increasing use of Artificial Intelligence (AI) related methods. The advent of Deep Learning (DL) and increasingly high performance neural networks – particularly Convolutional Neural Networks (CNN) – in computer vision tasks created promising applications for clinical and diagnostic domain. This is true for both, classification and segmentation, tasks related to medical datasets [1] [2] [3] [4].

More recently, the performance of human pathologists has sometimes been surpassed by these networks, such as in the task of detecting metastases in patches extracted from whole slide histopathology images (WSI) [4] [5].

Deep Learning architectures achieved very high levels of accuracy in these tasks, offering promising solutions for multiple issues pertaining to this problem. For instance, pathology experts performing the same task can disagree in more than 20% of the cases [6]. The related literature also mentions this type of task as tedious and time consuming, increasing the chance of human error.

These AI systems could greatly aid in balancing work loads of pathologists, who often are in great demand, and allow for more accurate and consistent diagnosis.

However, in general, a key issue within these promising results is that there is no clear indication of what makes a particular DL system, such as a CNN, output a certain prediction or classification for a given patch. Several authors cite this black-box behavior as a central problem of deep learning systems concerning medical data. The recent related literature calls for a solution to this issue, before AI can truly be adopted for this task in the medical world [3] [7] [8].

Since around 2016, Explainable AI (XAI) has been an increasingly blooming topic of research. A number of original papers and systematic reviews have been focusing on the development of techniques to explain decisions taken by AI models [7].

On the realm of image classification problems, multiple XAI techniques have been developed using different ideas. Many of these approaches involve backpropagation of the output to the input neurons, with specific functions or other specific operations conceived specifically for CNNs. Alber et al. provide a summary of these methods [9].

In this context, explanations consist in heat maps where each pixel in the image has a value according to its relevance for a given classification.

Another example is the Local Interpretable Model-Agnostic Explanations (LIME), developed by Ribeiro et al. [8]. In this approach, for image explanations, a given image is initially divided into super-pixels (also termed 'segments') and then their relevance, for a given classification, is determined using a linear model. The idea is that this simpler, and more easily interpretable, linear model could learn how to approximate the behavior of a more complex, non-explainable, non-linear

This work was supported in part by the Conselho Nacional de Desenvolvimento Científico e Tecnológico (CNPq) and Fundação de Amparo à Pesquisa do Rio de Janeiro (FAPERJ) under the process E-26/200.816/2019.

P.S. Iam, is with the Electrical Engineering Department, Pontifical Catholic University of Rio de Janeiro. (e-mail: iam.palat@gmail.com).

M.R.V. Marley, is with the Electrical Engineering Department, Pontifical Catholic University of Rio de Janeiro. (e-mail: marley@ele.puc-rio.br).

C.S. Eduardo is with the Electrical Engineering Department, Pontifical Catholic University of Rio de Janeiro. (e-mail: edusilva@ele.puc-rio.br).

model – such as a CNN.

The LIME algorithm is called model-agnostic because it only requires the outputs of the classifier for a given input image. In fact, LIME can be used for any image classification system, not just neural networks, as it doesn't employ any specific backpropagation procedures or any specific steps related to particular model architectures. Implementations of these approaches, most often in Python, have become available recently [8] [9].

Although scientific literature clearly states the need for XAI studies in medical imaging, there aren't many published works applying the existing approaches to medical imaging data.

In 2019, Palatnik-de-Sousa et al. [10] presented pioneering results using LIME to generate explanations for a WSI patch classifier trained on the Patch Camelyon dataset (P-CAM) [11], for detection of lymph node metastases. The choice of LIME over other techniques was justified both due to issues regarding the reliability of saliency based techniques [12], as well as the non-specificity of LIME when it comes to particular architectures. The model-agnostic nature of this technique meant it can be easily applied to any other type of architecture or classification model, unlike gradient based techniques that are more specific to Convolutional Neural Networks [9]. Considering these aspects, LIME could allow for more general comparisons to future studies.

Furthermore, the results achieved in [10] demonstrated the high influence of different segmentation methods on the generation of LIME explanations, for that particular dataset. Besides, the results also highlight that Felzenszwalb's segmentation algorithm [13] generates high explanation weight heat maps. These explanations agreed with medical expert segmentations for the same patches.

However, one difficulty was the necessity to choose segmentation parameters for generating super-pixels. Using default segmentation parameters within the LIME function might not generate meaningful explanations, and deciding the ideal set of parameters can be challenging without defining specific metrics for evaluating the heat-maps.

One of the alternatives proposed in [10] was the use of simplified, parameter-less, square grids to generate very rough simplified explanations. Despite showing that these "squaregrid explanations" in general agree with the more sophisticated segmentation algorithms, it was also noticed that a lot of finer detail is lost.

Considering these issues, the present study aims to explore a novel solution to this problem. In this way, the main goal of this paper is to present a more reliable and reproducible explainable model for explaining image classifiers. This methodology is based on multi-objective genetic algorithms, allowing for an automation of the parameter tuning process in image classification explanations. As this method/model entails using an evolutionary approach, it will be termed as "Evolved Explanations" (EvEx) throughout the manuscript.

By developing this methodology, the hope is to create a reliable way to peer into the black-box decision making process of convolutional neural networks, which are the most common type of classifier applied for medical imaging classification problems.

To demonstrate the performance of this new model, the manually optimized results obtained by Palatnik-de-sousa et al. [10] are compared to EvEx, using a previously studied CNN trained on the Patch Camelyon (P-CAM) dataset [11]. The idea is to show that the EvEx model can eliminate the necessity of manual parameter tuning for super-pixel creation. By defining metrics to evaluate the quality of explanations, and by exploiting the observed behaviors of Felzenszwalb's segmentation algorithm for this dataset [10], the generation of explanations can be automated, creating more detailed explanations than with squaregrid.

Additional contributions described along this paper involve studies about the reproducibility of the generated explanations. Namely, LIME is a technique with stochastic components, which can cause explanations to vary slightly in different runs for the same image. This study presents two approaches to counteract this potential issue, showing that it is possible to generate high quality, reproducible explanations, without requiring human expert fine tuning of parameters or the trade-off between parameter setting complexity and explanation details, like with squaregrid, a parameter-less method.

Section II contains a description of the Evolved explanation model developed in this study, highlighting the different parts that compose it. Section III describes the application of EvEx to the case study of lymph node metastases classification, to improve previously existing results. Section IV presents the achieved results and discussions of this approach, and, finally, Section V is the conclusion for this manuscript.

## II. The EvEx Model

In this section the EvEx model is described, and each one of its main components is detailed. Fig 1. shows a block diagram of EvEx, with the colored arrows representing the input/output relationship between each component.

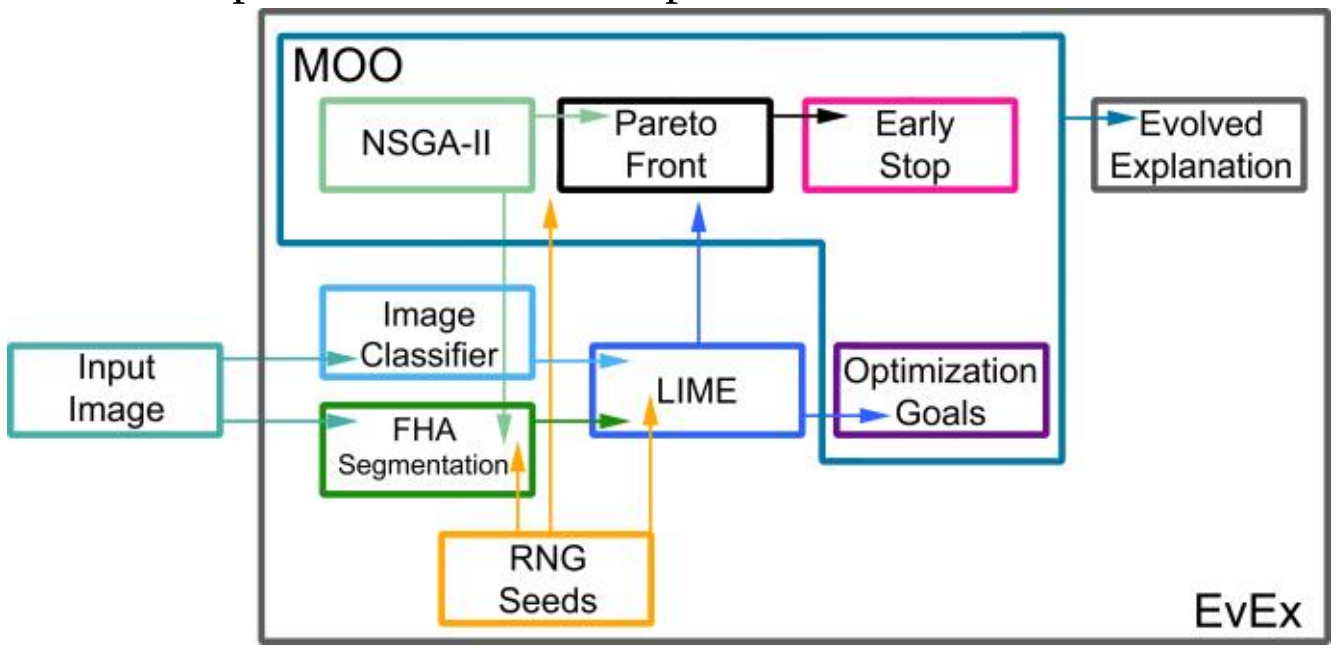

Fig. 1. Block Diagram of the proposed Evolved Explanations (EvEx) model. Each box shows a component of the model and the arrows indicate the input/output relationships among them.

As a brief overview, an input image passes through an image classifier. This image also undergoes a segmentation step, and the LIME methodology is applied to generate explanations. From these explanations a series of goal metrics are calculated. A Genetic Algorithm (GA) repeats this process a number of times generating multiple sets of segmentation parameters and creating a Pareto Front of the best explanations, which are averaged onto a final explanation, once the evolution process is over. An early stop component can control the duration of the GA process, in case no new Pareto Fronts are found after a given number of generations. Random Number Generator (RNG) seeds are used to evaluate reproducibility, and allow a proper comparison among the obtained results.

In the following subsections EvEx components, as well as the process of applying them, are described in more detail.

### *A. LIME*

LIME explanations [8] are generated by first dividing a given image into super-pixels or segments, using a segmentation algorithm. Super-pixels/segments are groups of pixels with similar colors, textures or other characteristics that hold some form of contextual information of a given image region. In section II.B, the choice of the segmentation algorithm for this study is discussed.

An example of this procedure can be seen in Fig. 2. Panel (a) shows a sample image. Panel (b) shows the super-pixels generated after applying a segmentation algorithm (see section II.B). For illustrative purposes only, these super-pixels are highlighted in yellow. However, these yellow markings are not applied to the real LIME implementation.

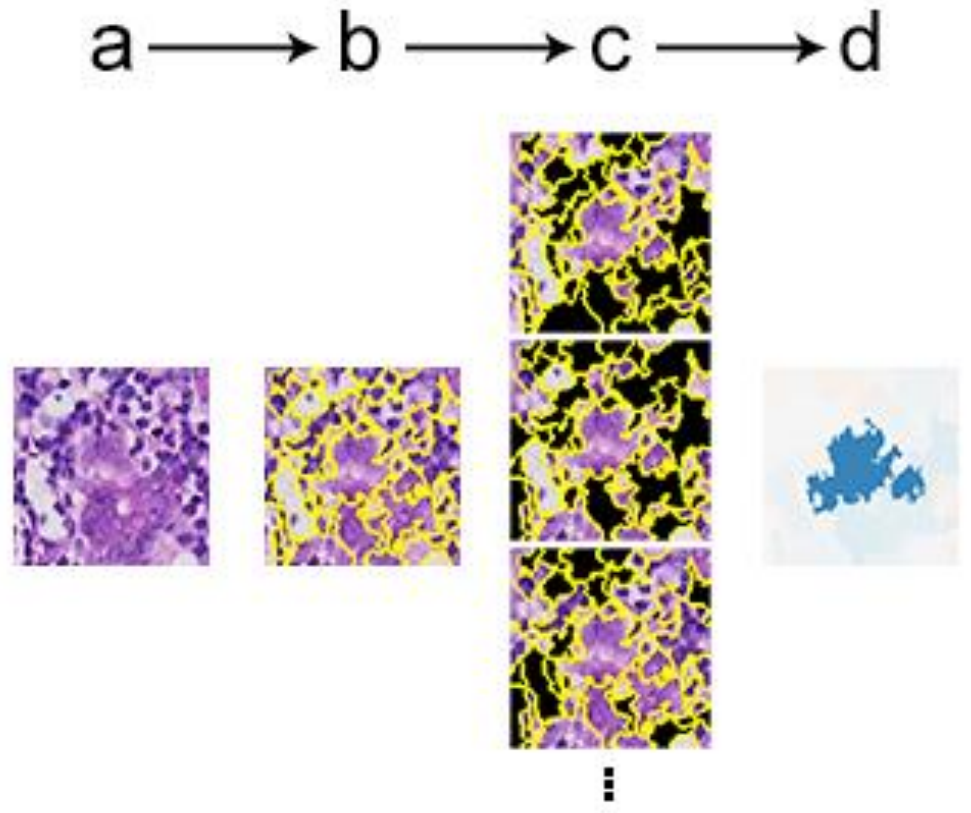

Fig. 2. Basic LIME diagram. Panel (a) shows a patch from the P-CAM dataset, correctly classified by the CNN model as class 1. Panel (b) shows the super-pixels found by a segmentation algorithm. The super-pixels are then covered in black at random (panel (c)) creating a distribution of perturbed images. From this perturbed distribution, a linear model learns to approximate the CNN prediction for this image, and the explanatory weight of each super pixel, which are then plotted as a red-blue heat map (panel (d)).

The segments are then randomly covered in black, creating a number of perturbed images, as shown in panel (c). Sections II.D.4 and 5, further below, contain a more detailed discussion on the number of perturbed images used in this case.

These images are then presented to the CNN model (see section II.C) being studied, that computes the prediction probabilities. With the perturbed images and their respective prediction probabilities, a linear model is trained, and the explanation weights (*xw*) for each super-pixel are then plotted in a blue-red heat map – panel (d), Fig. 2.

Color intensity is proportional to the absolute value of the weight. As such, blue regions indicate super-pixels that contribute towards the correct classification, and red indicate the opposite. This LIME implementation allows the user to generate explanations for each class separately, if requested. For a more in-depth discussion of the particularities of LIME see [8] [10].

### *B. Segmentation Algorithm*

Felzenszwalb's efficient graph based image segmentation (FHA) [13] generates an over-segmentation of an RGB image, using tree based clustering. The current implementation of LIME uses the FHA function from the scikit-image Python library [14]. Previously published results, by Palatnik-de-Sousa et al. [10], suggest that explanations generated with FHA typically result in at least one large super-pixel with high explanation weights, especially when compared to two other segmentation algorithms that were also evaluated with LIME: Simple Linear Iterative Clustering (SLIC) [15] and Quickshift [16]. For this reason, FHA was chosen as the segmentation algorithm for this study.

The idea is that FHA segmentations can be used to simultaneously maximize the explanation weight (*xw*) of the most relevant superpixel, while minimizing its area, if a Multi Objective Optimization (MOO) is used. By defining such objectives, the segmentation process could be automated, eliminating the need for manual parameter tuning, while, at the same time, highlighting the most relevant areas of the image, with finer detail than previously obtained by the manually tuned FHA. Essentially, these optimized explanations would allow for a more nuanced understanding of the image classifications.

Besides the already mentioned desirable behaviors of FHA for the purpose of multi-objective optimization, this algorithm also has two other marked advantages compared to SLIC and Quickshift, reinforcing it as an ideal candidate for this study. First, at least in the scikit-image implementation, FHA is the fastest segmentation algorithm. Secondly, it has less tunable parameters than either SLIC or Quickshift, greatly reducing the search space.

When combined, these factors contribute to massively reducing the computational cost of optimization – saving hours of runtime for each optimization. However, despite this being extremely relevant given the limited available computational resources, the main factor for prioritizing FHA is the desirable segmentation results reached for this algorithm, which returns large super-pixels with high explanation weights.

The FHA function parameters are:

- *Scale*: Indirectly controls the number of segments produced. In general, a larger value leads to larger segments. More specifically, as per [13], it sets a scale of observation for the threshold function, when calculating the minimum internal difference between components;
- *Sigma*: Standard deviation of the Gaussian Kernel used for blurring in preprocessing;
- *Min_size*: Minimum segment size, enforced in postprocessing.

### C. Image Classifier

This study uses a publicly available model trained on P-CAM for a Kaggle competition [17], also used in [10]. It is a convolutional neural network, summarized in Fig. 3.

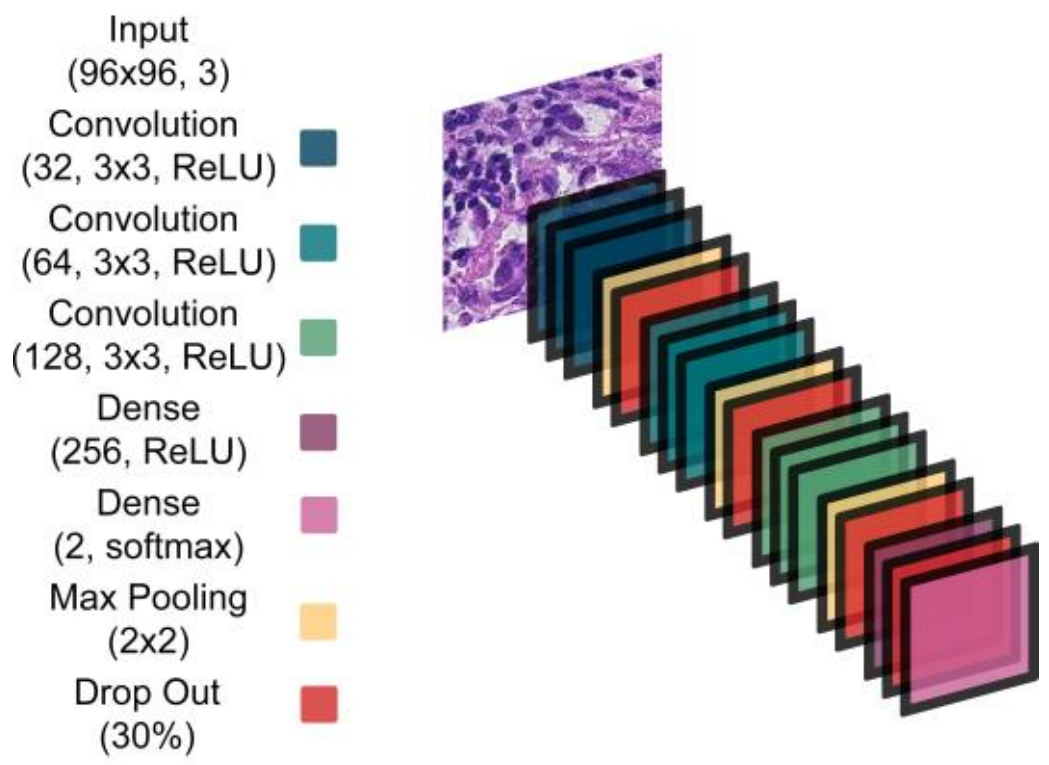

Fig. 3. CNN model used. Layers are described by the color code indicated in the left. The number of filters, size of filter and activation function are specified for the convolution layers. For Dense layers, the number of neurons and activation function are highlighted. For the max-pooling layer, the size of the pooling element is shown. For the drop out layer, the dropout percentage is presented.

### D. Multi-Objective Optimization

As discussed in subsection II.B, FHA segmentation is a promising candidate for automatically adjusting parameters and generating high quality explanations.

For the case of FHA segmentation, each individual of the developed GA has 3 genes, representing the FHA parameters: scale, sigma and min_size. The evolutionary process seeks to find the best set of parameters, responsible for generating the best segmentation as well as the best explanation heat-maps.

#### 1) Optimization Goals

Considering the size of the search space and complexity of the evaluation function, a genetic algorithm was selected to perform the pursued multi-objective optimization. As such, a non-dominated sorting genetic algorithm function (NSGA-II) [18] was used. This MOO aims to optimize three objectives:

- Explanation score (*Es*): LIME provides this quantity as a value between 0 and 1. It corresponds to the $R^2$ of the ridge regression model, created on the last step of LIME;
- Largest weight (*Lw*): Once the explanation is generated, each super-pixel has an explanation weight. *Lw* is the largest weight among all super-pixels. In other words, it is the explanatory weight of the most relevant super-pixel. It can range from 0 to 1, with typical values being closer to 0.7~0.9;
- Relative area of most relevant segment (*Ar*): This is the relative area covered by the most relevant super-pixel. It is computed by dividing the number of pixels of this segment, by the overall number of pixels in the image. The choice of relative area, instead of just the area, is so that this quantity also ranges from 0 to 1, as the other optimization parameters. This is a key metric to explore the desirable behavior where FHA tends to find large segments that dominate the explanation.

With these goals, the NSGA-II algorithm is set up to maximize the Explanation score (Es) and Largest weight (*Lw*), while minimizing the Relative area of most relevant segment (*Ar*). In other words, the NSGA-II tries to find the highest scoring explanations, with the highest weighted most relevant super-pixels, while at the same time minimizing the chance of explanations that just cover the entire image in one single giant super-pixel.

As a result of the MOO, a Pareto front is generated, with one or multiple explanations that can be considered equally optimized. The final explanation given by the MOO is the average of these Pareto front explanations.

To justify why this average is used, one must imagine the process of trying to obtain a single best ideal explanation. This would mean optimizing the explanation down to effects caused by single pixels, which is not realistic, even from a medical standpoint. Thereby, the implemented MOO is less concerned with finding a perfect unique explanation down to single pixel precision, and more with finding the consensus between plausible explanations of the highest quality solutions achievable with the given goals.

#### 2) Hyper-volume

If each solution for an MOO problem is considered a point in an n-dimensional space, a common indicator of MOO performance is the n-dimensional volume contained within the solution set [19]. That is to say, the space contained within the solution points. This metric is called Hyper-volume (HV).

Since the MOO here involves 3 objectives, the HV is the three dimensional volume contained within the solution sets, which in this case are the Pareto fronts.

To keep track of the GA performance, this HV is calculated for the Pareto fronts found in each generation. The HV computation function described by Fonseca et al. [20] was used in this work.

This HV function, however, implicitly assumes that the MOO aims to minimize all goals. For the *Ar* goal this doesn't change anything, since it was already a minimization problem. But *Es* and *Lw* should be maximized. Therefore, the MOO and HV functions consider minimizing (1-*Es*), (1-*Lw*), and *Ar* instead, respectively.

The HV is then computed using the (1,1,1) point as reference, meaning that (1,1,1) would correspond to the worst possible solution, and (0,0,0) to the best, for all goals.

This hyper-volume is just used to observe how the Pareto front is evolving. It is not used to interfere or influence the evolution process in any way. The expected result is to see the HV value increasing as the evolution process occurs and the Pareto front grows away from the (1,1,1) point towards the neighborhood of (0,0,0).

### 3) Early Stop

During the first tests with EvEx, the GAs were set to run for a number of generations (in this case 200), but seemed to converge before that. In order to reduce computation times, an Early Stopping criterion was adopted.

For each generation, the current Pareto front is saved into a list. If no new Pareto fronts are found, compared to the ones already existing on this list, a control variable is incremented by 1, starting from 0. If this control variable reaches a chosen value (in the case of this study, 70), the run is terminated.

### 4) Reproducibility

One of the key aspects of performing this analysis is the fact that LIME explanations have some inherent variability. That is to say, generating multiple LIME explanations for the same image can create slightly different explanations. This happens because of the random generation of perturbed images, within LIME. Even if the same segmentation algorithm is always used, the resulting explanation weights, for each segment, can vary depending on which super-pixels are covered in the randomly generated perturbed images, as depicted in Fig.1.

This could be a problem for the genetic algorithm, since it implies that each individual in the GA population does not have a unique evaluation. Therefore, it could end with multiple identical individuals with slightly different evaluations. To counter this, two approaches were evaluated and added to the EvEx model as a reproducibility component.

The first one consisted in simply increasing the number of perturbed images used in LIME, which decreases variability in the goal metrics, but can't eliminate it completely. This approach still returned populations that had identical individuals with different evaluations, albeit this difference becomes almost negligible.

The other approach is to set randomizer seeds, so that the LIME explanations will always use the same perturbed image distributions. This makes the outcomes consistent when LIME is called multiple times on a given image. However, on a different seed the results could be somewhat different.

In practice, a combination of both approaches was used. Multiple runs were performed, each with different random seeds, so that within each run the LIME results are consistent, but they can be compared between different runs to see how the explanations behave. Simultaneously, the number of perturbed images was tuned so as it is low enough to allow for fast computation, while sufficiently high to still decrease variability to within a controlled number of decimal places, for the goal metrics. Such differences were computed as standard deviation heat maps, described on the next sections.

In other words, the same individual would always have the same evaluation within a given run, and would still have similar evaluations between different runs. Setting the number of perturbed images in 200 demonstrated to be enough.

As discussed further below in the results, this combined approach generated explanations that largely reach the same consensus between the Pareto fronts evolved.

More importantly, this makes the generation of explanations fully deterministic for a given set of parameters and random seed, which is crucial for the reproducibility of explanations.

### 5) Comparing Heat Maps

In order to compare the explanation results of each seed and analyze how much they may vary, each patch had its pixel-wise standard deviation (SD) computed over the four explanations/heat maps, corresponding to the four different seeds studied. Similarly, the pixel-wise means were computed, and subsequently its absolute values.

Then, the pixel-wise relative standard deviation (RSD) (sometimes also called coefficient of variation) was computed by dividing the above described quantities. This quantity is plotted as a grayscale heat map, with a color scale ranging from 0 to 1, as an RSD value greater than or equal to 1 serves as indicator of high variance.

An explanation score threshold was set. Pixels that had mean explanation scores below this threshold were excluded from the RSD heat map, because regions with low explanation scores are not meaningful to interpret a classification.

This last step is further justified since, as seen in [10], regions of low explanation scores may change a lot between different LIME runs. On the other hand, regions of high explanation scores, which are the most relevant ones to describe the classifications, typically fluctuate much less.

## III. Case Study: Lymph Node Metastases

### A. Dataset

The CNN model used in this manuscript and [10] was trained on the Patch Camelyon (P-CAM) dataset, derived by Veeling et. al [4] from the Camelyon 16 hematoxylin and eosin stained WSIs.

Each 96 by 96 patch has binary labelling that indicates the presence (label 1) or absence (label 0) of at least one pixel of tumor tissue in the center of the patch – a 32 by 32 pixel square. The dataset has a class balancing close to 50/50.

Although the dataset was originally made available on a github repository [21], the version used here is the one available on the Kaggle website [11]. The latter is similar to the github version, except for the removal of duplicate patches caused by probabilistic sampling. Throughout this manuscript, whenever the P-Cam dataset or patches are mentioned, they refer to the Kaggle version.

### 1) Medical Annotation

The WSI Camelyon16 dataset includes manual annotations of which parts of the images are metastatic tissue. This annotation/segmentation was done and verified by students and expert pathologists at two different Dutch hospitals (Radboud University Medical Center and University Medical Center Utrecht) [6].

Mappings of these WSI annotations to the patches were made available by Veeling et al. on the PCAM github repository [21]. In [10] they were then mapped onto the Kaggle version, and these mappings are once again used here.

#### 2) Patch Selection

Considering the results achieved in [10], patches classified as true positives, where the medical annotation is a sub-region of the patch, can be a helpful option to evaluate the methodology here presented.

The key reason is that, in these patches, it is immediately apparent if the evolved explanations agree or not with the medical segmentation, as seen in Fig.5 panel (a), for instance.

### B. LIME Parameters

Like in [10], the original LIME implementation in Python was used [8]. For the most part, the LIME parameters used in this work were similar to the previous study [11]. The main difference is the number of perturbed images, which was decreased for this study as previously explained in section II.

### C. FHA Parameters

Although manual tuning has allowed for results that match medical expert segmentations [10], the manual search for the ideal set of parameters for the FHA can be challenging and considerably time consuming. Comparing heat maps visually with no defined metrics can also often be non-trivial. As discussed in the previous section, this motivates the use of a multi-objective optimization. The search space for the FHA parameters is described in more detail on the next subsection.

### D. Genetic Algorithm Parameters

The DEAP Python library [22] was used for the GA component of EvEx. The basic parameters used in the developed genetic algorithm are as follows:

- Population size: 80
- Maximum number of generations: 200
- Mutation operator probability (*mutpb*): 20%
- Crossover operator probability (*cxpb*): 50%

The search space is defined considering the three FHA parameters (scale, sigma and min_size). An individual in the population corresponds to a set of these three parameters, defined as:

- Scale: 3 decimal float, range: [1,1000]
- Sigma: 2 decimal float, range: [0, 5]
- Min_size: int, range: [15,500]

The lower and upper limits for these ranges were chosen based on the tests ran in [10], as well as in preliminary tests ran for this manuscript. Values outside these ranges don't seem to generate useful explanations, and increase the size of the search space needlessly. Specifically, regarding the min_size parameter, values around 15 and below tend to enable segmentations with an exceedingly large number of small segments, which not only increase computation times for LIME, but also do not generate good explanations. Previous results presented in [11] also indicate that generating a very large number of small segments doesn't allow them to individually hold much relevant explanatory information [10].

Regarding mutation and crossover, DEAP uses two sets of probability parameters. The first set, already mentioned above (*mutpb* and *cxpb*), refers to the probabilities that the specified mutation and crossover operators will be applied.

Each of these operators however may consist on standard functions of the DEAP library, or custom ones defined by the user. They may have individual internal probabilities (*indpb_mutation* and *indpb_crossover*) specific to the type of calculations performed on individuals.

The crossover function used in this study was the DEAP uniform crossover (*cxUniform*), with an *indpb_crossover* of 20%. This function was chosen over one-point and two-point crossover, because preliminary tests performed with these other functions seemed to show premature convergence of the solutions, with Pareto fronts not changing anymore after the first generations. The use of uniform crossover successfully corrected this behavior.

As for the mutation parameters, each gene in the individuals had a specific mutation function associated with it. Gaussian mutations were used for the FHA scale parameter (Mu = 0, sigma =10), as well as the FHA sigma parameter (Mu = 0, sigma = 0.05). Uniform integer mutations were used for the integer parameter (min_size). The int range for the mutation was [15,500] The value of *indp_mutation* was set to 20%.

## IV. Results and Discussion

### A. Pareto Fronts and Hyper-volume

Fig. 4 shows an example of Pareto front and HV behavior from a given evolved explanation. More specifically, it is the generated result for the image highlighted in Fig. 5 row (a), seed 45. The initial population and final Pareto front evaluations are plotted in panel (a). The shape of the optimized Pareto front becomes apparent, closer to the optimal (0,0,0) point.

Indeed, most of the initial 80 individuals of the starting population, in orange, are clumped in two groups. The first one is composed by individuals around the (1,1,1) point of the plot, which correspond to segmentations where the most explanatory super-pixel covers a large area of the image, but fails to explain the classification, once they have low explanation score and explanation weight.

On the other hand, the second clump of individuals are grouped close to the origin, upwards along the Z-axis. This second clump corresponds to the individuals that have large explanation scores and weights, by sheer virtue of having the most explanatory super-pixel covering a large area of the image, once the relative areas lie between 0,3 and 1,0. Indeed, this seems to be the majority of cases. As previously discussed, this is caused by the FHA segmentation, and this behavior is what the optimization seeks to exploit.

Panel (b) shows that the hyper-volume of the Pareto front increases throughout the run, demonstrating the algorithm is able to find individuals optimizing the defined goals.

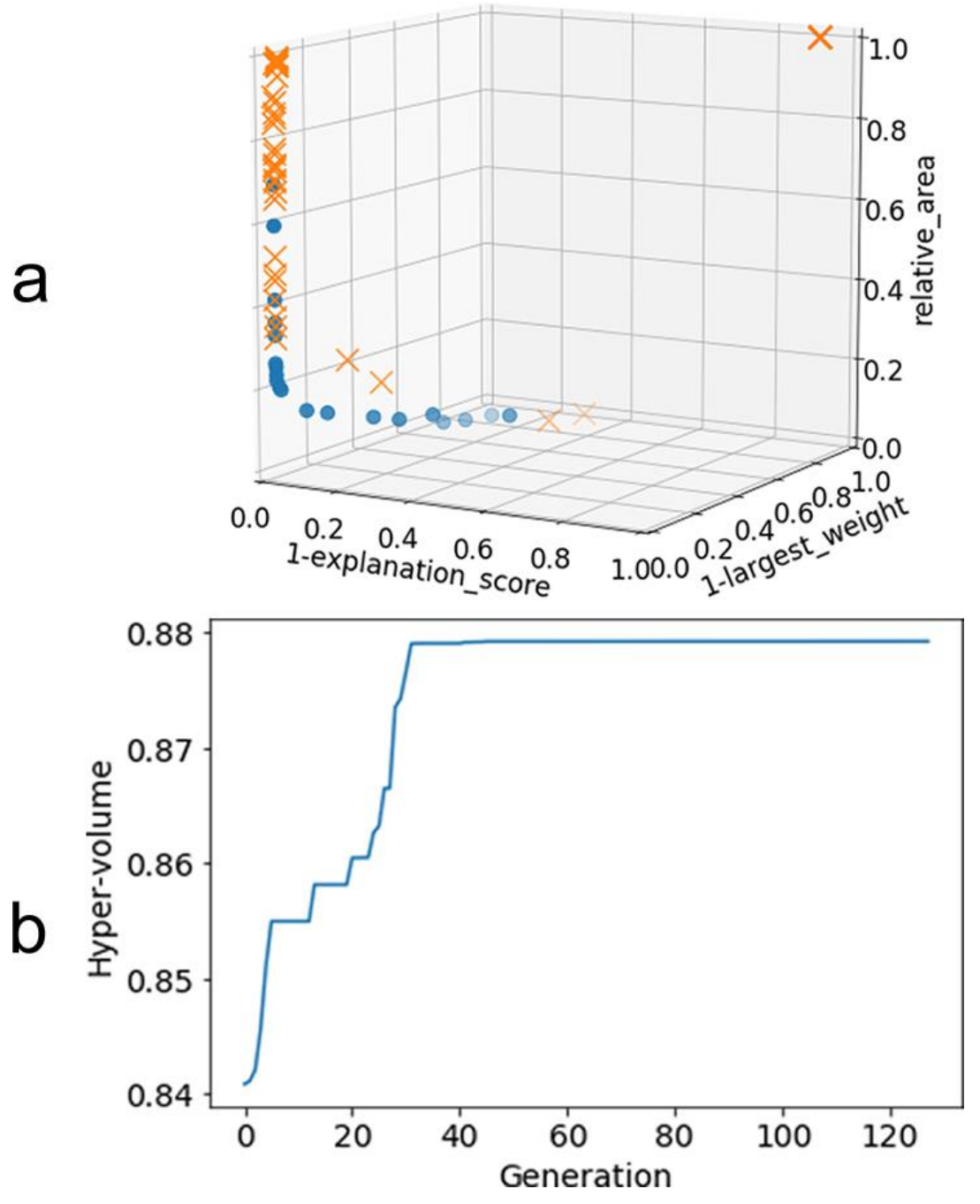

Fig. 4. Pareto front and hyper-volume plot. Panel (a) shows the initial population (in orange) and final Pareto Front (in blue) for a given sample run (the patch corresponding to Fig. 5, panel (a), seed 45). The plot delineates the shape of the Pareto Front after the last generation of the GA. Panel (b) shows how the hyper-volume metric behaves throughout the evolution process.

When comparing the Pareto front of the initial population with the one achieved for the final population, it is clear that low performing individuals progressively were replaced in the populations by better counterparts. The final front includes individuals with super-pixels covering much smaller regions of the image, while still maintaining high explanation and score metrics all around.

### B. *Evolved Explanations*

In total, considering all images studied, and all seeds, there were 32 runs of EvEx. The average size of the final Pareto fronts was 37 individuals. The largest front observed had 61 individuals and the smallest 15. Fig. 5 shows the outputs of these 32 runs. These outputs consist on the averaged explanation generated with the individuals from the final Pareto front, after the last generation of the GA.

For the cases in rows (a), (b), (c), (d), (f), (g), (h), the most highlighted area of the image (deepest blue, corresponding to highest explanation weights) is contained within the medical segmentation (displayed as a green transparent overlay). The image on row (e) has lower explanation weight areas scattered across the entire patch, but the deepest blue regions are still mostly contained within the medical segmentation, extending outwards of it to the top left of the patch.

Because of the deterministic approach adopted in the methodology for this study, results for each seed are exactly reproducible on different runs. However, even comparing the results from different seeds, they also seem to indicate agreement as to which areas of the images are most relevant.

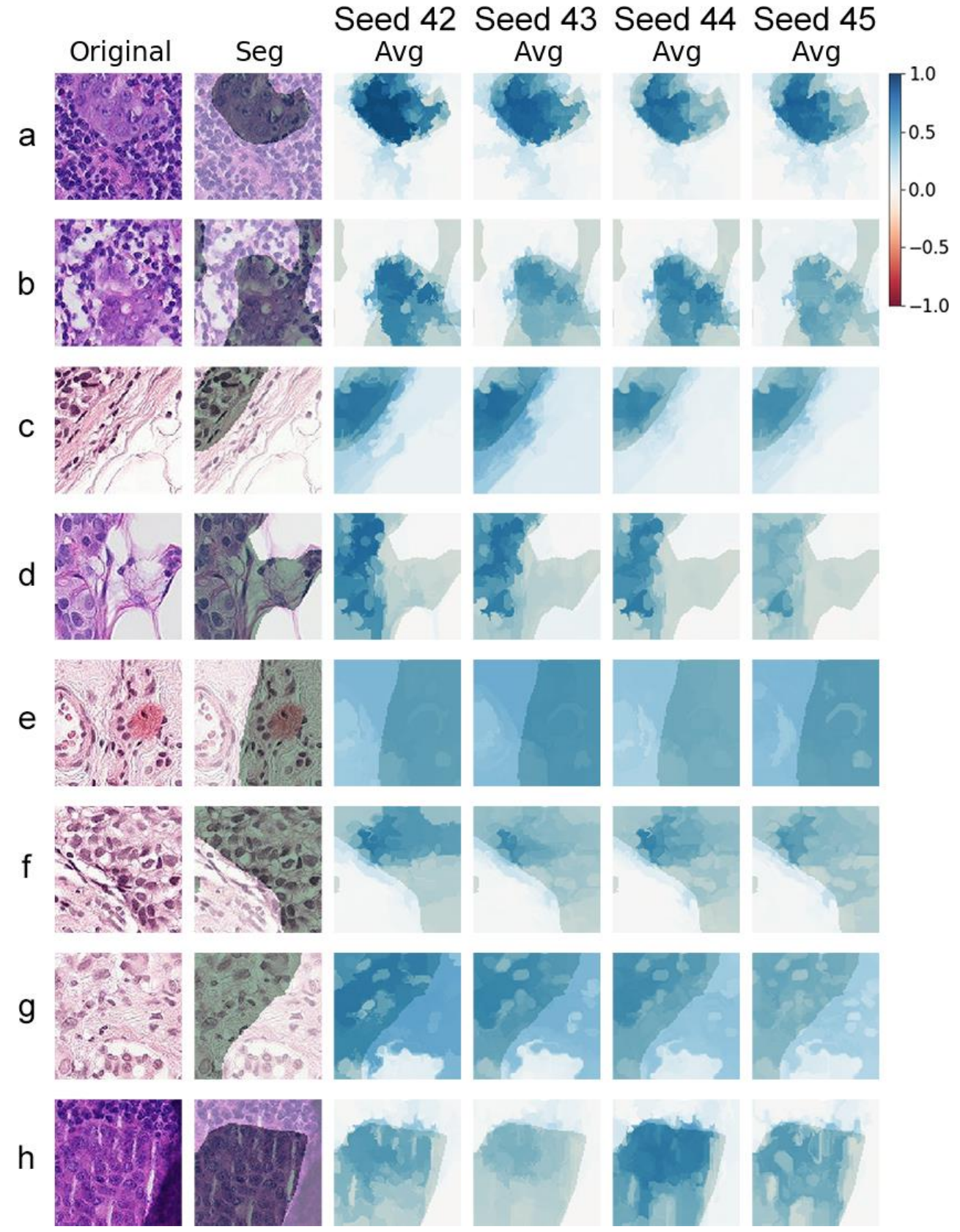

Fig. 5. Rows (a)-(h): Evolved explanations for seeds 42 through 45. ‘Avg’ represents the average of the final Pareto Front explanations achieved at the end of the evolution process. A Blue-Red color scale is used to represent explanation weights. ‘Seg’ is the medical expert segmentation for that given image, represented by a green overlay.

This happens because each final Pareto Front might have a different number of individuals, focusing on different parts of the images. However, it seems that aggregating those individual explanations led to consistent averaged explanations. This averaging might also lead to robustness against the randomness expected for LIME explanations, making this technique a better candidate for such applications.

### C. *Variability and Reproducibility*

To further test and quantify this agreement between the seeds, a second set of plots analyzing the relative standard deviation (RSD) between images was generated, as described on section II.D.5. Fig. 6 shows these results.

The standard deviation (SD) column shows the pixel-wise standard deviation computed between the 4 seeds, for each image. The maximum SD occurs in a sub region of image (h); however, it comprises very few pixels of the image. For most cases, very light colors are observed throughout the SD heat maps, meaning the standard deviations assume low values.

However, it is hard to draw conclusions from just this quantity alone, as different pixels have different mean explanation weights on the original blue-red heat maps. A standard deviation of 0.1, for instance, is drastically more expressive for a pixel with mean explanation weight of 0.12, than for another one with 0.9.

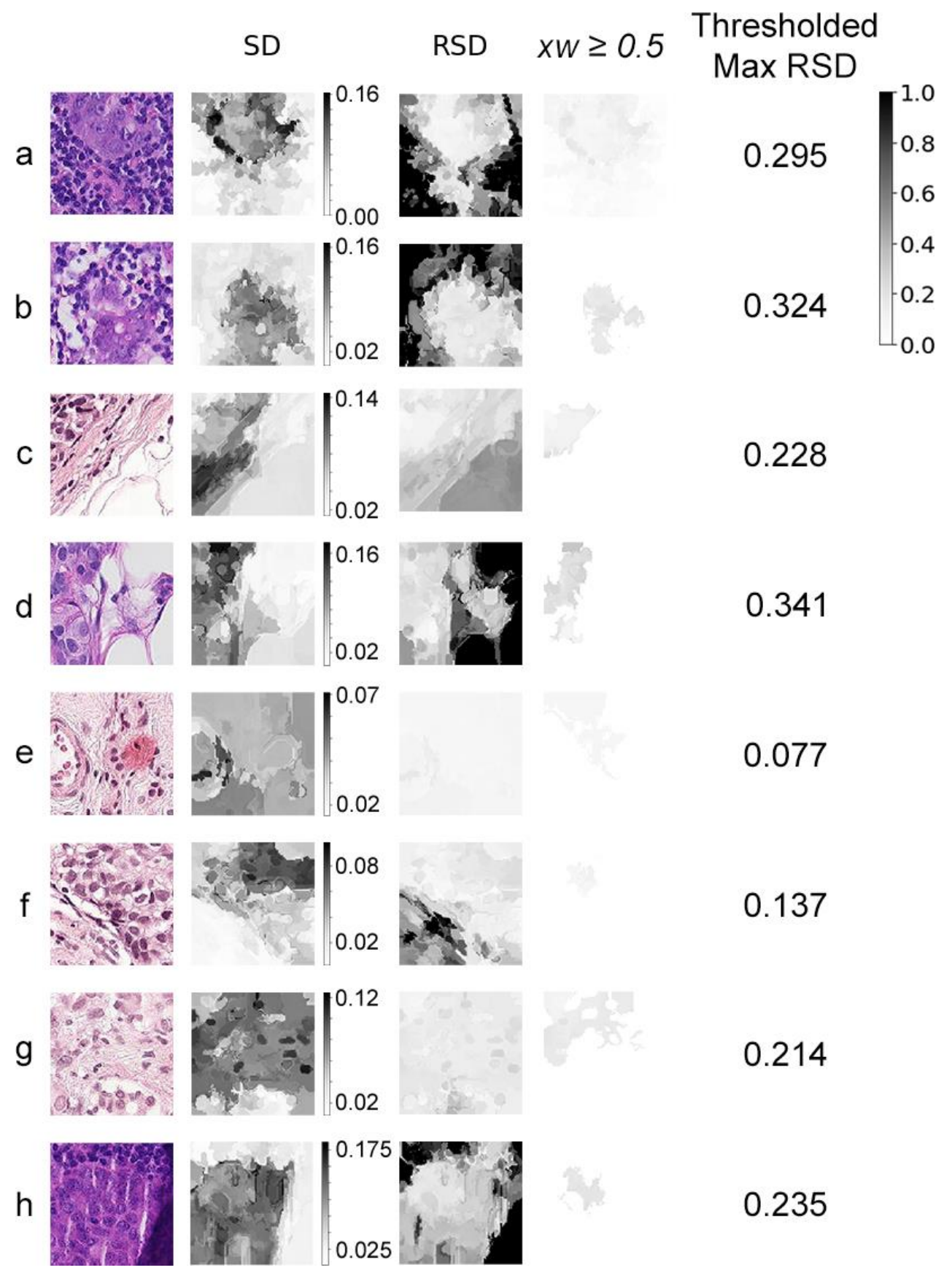

Fig. 6. Rows (a)-(h): Pixel-wise Standard Deviation (SD) and Relative Standard Deviation (RSD), for the 4 seeds studied. The RSD and threshold RSD use the color scale on the top right. The SD column uses individual color scales presented to the right of their corresponding heat maps.

To better visualize this effect, as explained in section II.D.5, RSD heat maps were also generated, using a common grayscale from 0 to 1 for all analyzed cases. In this scale, values equal or greater than 1 indicate that the standard deviation is equal or greater than the mean, indicating high variability. Then, the scale is capped at that value. Consequently, points with RSD > 1 will show up as black on those heat maps.

Taking Fig. 6 (a) as an example, it can be immediately noticed that there are in fact regions with RSD of 1 or more. However, comparing Figs. 6 and 7, it becomes apparent that these high variability regions are areas of extremely low, negligible explanation weights. This result is, in fact, in line with previous observations by Palatnik-de-sousa et al. [10] that low explanation weight regions tend to vary more between LIME runs, most likely because they are not relevant to the explanation. Although their weights fluctuate, they do so in very small absolute values, with many decimal places. Thus, it is both expected for these areas to be highly variable in regards to RSD, and also to be highly irrelevant for the explanation. Therefore, they can be safely disregarded from the explanation. This pattern is, in fact, observed for most of the images studied, although Fig. 6 (c), (e) and (g) seem to show high agreement (low RSD) for the entire patch.

More relevantly, as expected, the areas of high explanation weights (strong blue colors in Fig. 5) seem to coincide with areas of low RSD in Fig. 6, meaning that the explanations generated by applying this methodology, with different seeds, seem to agree in general as to what areas of the patch hold the most information to explain the classification. This agrees with the previous discussion; however the RSD analysis helps to quantify and solidify this interpretation.

To better visualize this behavior, a third set of RSD heat maps was generated, by using an *xw* threshold, that excludes from the RSD plot any pixels with mean explanation score (averaged between the 4 seeds) below this threshold.

The threshold used was 0.5, which overall is not a high explanation weight, considering most images displayed regions with *Es* above 0.6. But, even for this low threshold, there is remarkable agreement between different seeds, with the maximum RSDs observed being at most around 0.3, as highlighted in column 'thresholded maximum RSD' of Fig. 6. Furthermore, these maximum RSDs (stronger gray hues) are only observed in small sub-regions of the heat maps, with the majority of the areas having even lower RSD values.

Notably, taking an image such Fig. 6 (a), that originally had regions of high explanation weights, above 0.8, it is possible to vary the threshold and see how it affects the RSD. Namely, the maximum RSD of 0.295, at 0.5 *xw* threshold, could be considerably lowered to 0.073, at a 0.8 *xw* threshold. This behavior indicates that for a given patch, using different seeds, explanations generated with this methodology show agreement between areas with increasingly higher explanation weights. It also means that it seems reasonable to expect areas of high *xw*, above 0.7 or more, to agree considerably between different seeds. It is encouraging that even areas of moderately low *xw* seem to show RSD values much smaller than 1.

The *xw* threshold column of Fig. 6 also resembles, from a quantitative standpoint, the intuitive similarity between the heat maps generated from different seeds, for each original image, that might be noticed at looking at Fig. 5. The areas that seem to agree, in general, are indeed the ones that correspond to low RSD values.

As such, the methodology here employed seems to be fully deterministic within individual seeds, as well as agreeing considerably between different seeds. It makes this GA MOO approach a very robust, automated alternative to manual parameter tuning for segmentation and generation of explanations. Aiming to generate more explainable and reliable systems in medical AI settings, this seems to be a promising result. The very important factor of results reproducibility is also greatly enhanced with the proposed methodology, allowing for more easily traceable explanations.

If medical experts so desired, they could dissect a given explanation by taking a specific seed, viewing each of Pareto front individual explanations separately, for that seed, as well as varying *xw* thresholds on the heat-maps, varying the range of the color scales, among other such ideas that might aid in visualizing results. The more cumbersome, time consuming and non-trivial task of finding appropriate parameters for segmenting patches and generating explanations can be left for the multi-objective GA here presented.

### D. Further considerations

The main drawback of EvEx is its computational cost. Each 200 generation GA run took between 6 and 8 hours, to generate an explanation for one patch. However, upon analysis, it was shown that the factor that contributes the most for this high computational time is the LIME function. It is possible that more computationally efficient implementations of LIME can be developed in the future, or that this same methodology could be employed in much more powerful hardware than the Kaggle cloud kernels used in this study, allowing faster processing of the patches.

Akin to the previous results reported in [10], a notable feature of the explanations here obtained is that, starting from a very simple binary label on the P-CAM dataset, the explanations generated show heat maps rich in information, largely in agreement with medical expert segmentations.

However, a key difference is that when using manual parameter tuning, as in [11], one might use human expertise or expert segmentations to guide the decision of which explanations are best. In the case of this evolutionary algorithm, the only factors guiding the explanations are the goal metrics to be optimized. The medical segmentations are only used for comparison, after the evolution process is over. Yet they show remarkable agreement, as shown in Fig. 5.

This confirms the proposed technique as a valuable, reproducible methodology that can be further applied in other medical image datasets. This approach could prove to be extremely valuable, especially for cases where medical expert segmentations are not available, or in searching for explanations in medical imagining problems, where it is not yet known what regions of a given image should be relevant for classification.

## V. Conclusion

In this manuscript, EvEx, a novel XAI model that uses LIME explanations combined with a multi-objective genetic algorithm, was presented. The explanations generated are the averaged contributions of the individuals of the evolved Pareto front. The algorithm seeks to adjust the three parameters of an FHA segmentation function, which has the desirable trait of finding large super-pixels with high explanation weights. The goals defined attempt to capitalize on this property by simultaneously trying to maximize the explanation score and the largest weight presented by the super-pixels, while minimizing the area of the super-pixel with largest weight.

The EvEx model presented deterministic explanations for classifications of lymph node metastases, within random seeds. Additionally, it shows agreement between different seeds, especially in areas with high explanation weights.

These evolved explanations further agree with medical expert segmentations for the same images, while not requiring any expert manual tuning of LIME or segmentation parameters. Therefore, they provide an important step towards peering reliably into the black-box decision making of neural networks used for medical image classifications. Future projects may focus on improving computation times and applying EvEx to other types of image classification tasks outside of the medical area.

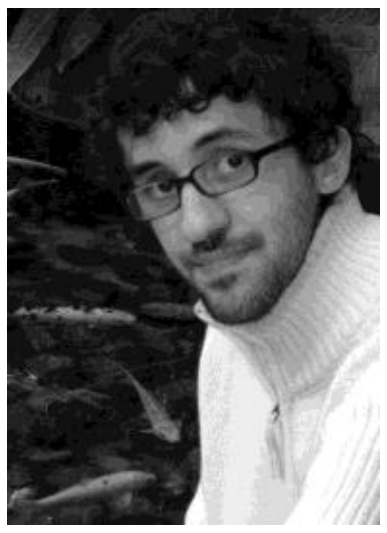

**Iam Palatnik-de-Sousa** received a B.Sc. degree in physics from the Federal University of Rio de Janeiro, and a M.Sc. degree in metrology from the Pontifical Catholic University of Rio de Janeiro.

From 2017 to the present he has been a PhD student on the electrical engineering department at the Pontifical Catholic University in Rio de Janeiro, developing a project related to explainable artificial intelligence for medical imaging. His previous research experience/interests include medical device safety from a metrological perspective; the study of fundamental properties of magnetic materials and molecular modelling and dynamics for the common prion protein.

Mr. Palatnik-de-Sousa has been indicated for the “Doutorado nota 10” scholarship by Foundation for Research Support of the State of Rio de Janeiro (FAPERJ).

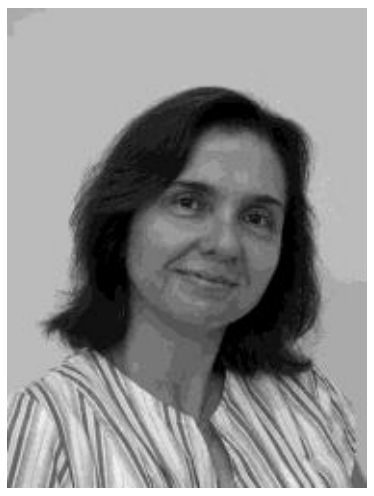

**Marley Maria B R Vellasco** received the BSc and MSc degrees in Electrical Engineering from the Pontifical Catholic University of Rio de Janeiro (PUC-Rio), Brazil, in 1984 and 1987, respectively, and the PhD degree in Computer Science from the University College London (UCL) in 1992. She is the founder and the Head of the Computational Intelligence and Robotics Laboratory (LIRA) at PUC-Rio.

Her research interests are related to Computational Intelligence methods and applications, including Neural Networks, Fuzzy Logic, Hybrid Intelligent Systems (Neuro-Fuzzy, Neuro-Evolutionary, and Fuzzy-Evolutionary models), Robotics and Intelligent Agents, applied to decision support systems, pattern classification, time-series forecasting, control, optimization, and Data Mining. Dr. Vellasco has served on the Board of Governors of the International Neural Network Society and is currently Vice-President for Conferences of the IEEE Computational Intelligence Society. She has supervised more than 40 Ph.D. Thesis and 85 MSc Dissertations and has coordinated more than 50 research projects with industries, some of them resulting in Technology innovation prizes. She is the author of four books and more than 450 scientific papers in the area of soft computing and machine learning.

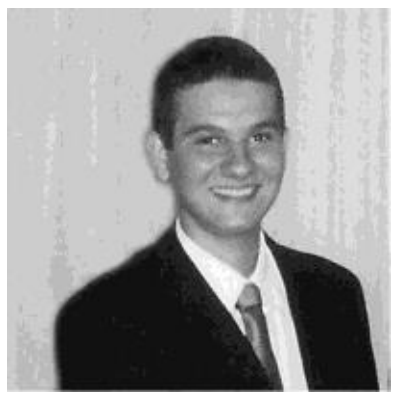

**Eduardo Costa da Silva** (M’08) received B.Sc. degrees in electrical engineering (Electronics and Process Control) and automation and control engineering from the Pontifical Catholic University of Rio de Janeiro (PUC-Rio), Brazil, in 2008. He received the M.Sc. in metrology (Quality and Innovation) and the D.Sc. in electrical engineering (Applied Electromagnetism) from PUC-Rio, in 2010 and 2013, respectively.

Since 2014, he has been a Professor with the Department of Electrical Engineering at PUC-Rio. He is the author of more than 50 articles and holds 4 patents. His research interests include electronic instrumentation, magnetic sensors, signal processing, biomagnetism, nondestructive evaluation and artificial intelligence.